\documentclass[conference]{IEEEtran}
\IEEEoverridecommandlockouts
\usepackage{cite}
\usepackage{amsmath,amssymb,amsfonts}
\usepackage{algorithmic}
\usepackage{graphicx}
\usepackage{textcomp}
\usepackage{xcolor}
\usepackage{booktabs}
\usepackage{subcaption}
\usepackage{placeins}
\usepackage{dblfloatfix}
\usepackage{tikz}
\usetikzlibrary{arrows.meta,positioning,calc}

\def\BibTeX{{\rm B\kern-.05em{\sc i\kern-.025em b}\kern-.08em
    T\kern-.1667em\lower.7ex\hbox{E}\kern-.125emX}}
\begin{document}

\title{Trajectory Generation with Endpoint Regulation and Momentum-Aware Dynamics for Visually Impaired Scenarios

\thanks{\textsuperscript{\textasteriskcentered} Corresponding author}
\thanks{This work was supported in part by the Fundamental Research Funds for the Central Universities under Grants ZYGX2025YGLH001 and Grant ZYGX2024XJ038, in part by the National Natural Science Foundation of China under Grants 62276055 and 62406062, in part by the Sichuan Science and Technology Program under Grants 2024NSFSC1476 and 2023YFG0288, in part by the Sichuan Provincial Major Science and Technology Project under Grant 2024ZDZX0012.}
}

\DeclareRobustCommand*{\IEEEauthorrefmark}[1]{%
  \raisebox{0pt}[0pt][0pt]{\textsuperscript{\footnotesize #1}}%
}

\author{
\IEEEauthorblockN{
Yuting Zeng\IEEEauthorrefmark{1},
Manping Fan\IEEEauthorrefmark{1}\textsuperscript{\textasteriskcentered},
You Zhou\IEEEauthorrefmark{1},
Yongbin Yu\IEEEauthorrefmark{1},
Zhiwen Zheng\IEEEauthorrefmark{1},
Jingtao Zhang\IEEEauthorrefmark{2},
Liyong Ren\IEEEauthorrefmark{1}\\ and
Zhenglin Yang\IEEEauthorrefmark{3,2}\textsuperscript{\textasteriskcentered}
}
\IEEEauthorblockA{
\IEEEauthorrefmark{1}School of Information and Software Engineering, University of Electronic Science and Technology of China,\\ Chengdu 610054, Sichuan, P.R. China.\\
\IEEEauthorrefmark{2}School of Medicine, University of Electronic Science and Technology of China, Chengdu, Sichuan, China.\\
\IEEEauthorrefmark{3}Sichuan Provincial Key Laboratory for Human Disease Gene Study, Sichuan Academy of Medical Sciences \& \\Sichuan Provincial People's Hospital, University of Electronic Science and Technology of China, Chengdu, Sichuan, China.\\
Email: \texttt{1030839769zengyuting@gmail.com}, \texttt{1244568901@qq.com}, \texttt{fmpfmp@uestc.edu.cn}, \texttt{ybyu@uestc.edu.cn},\\
\texttt{z.z.w@icloud.com}, \texttt{jtzhanguestc@gmail.com}, \texttt{lyren@uestc.edu.cn}, \texttt{yangzhenglin@cashq.ac.cn}
}
}

\maketitle

\begin{abstract}
Trajectory generation for visually impaired scenarios requires smooth and temporally consistent state in structured, low-speed dynamic environments.
However, traditional jerk-based heuristic trajectory sampling with independent segment generation and conventional smoothness penalties often lead to unstable terminal behavior and state discontinuities under frequent regenerating. This paper proposes a trajectory generation approach that integrates endpoint regulation to stabilize terminal states within each segment and momentum-aware dynamics to regularize the evolution of velocity and acceleration for segment consistency. Endpoint regulation is incorporated into trajectory sampling to stabilize terminal behavior, while a momentum-aware dynamics enforces consistent velocity and acceleration evolution across consecutive trajectory segments. Experimental results demonstrate reduced acceleration peaks and lower jerk levels with decreased dispersion, smoother velocity and acceleration profiles, more stable endpoint distributions, and fewer infeasible trajectory candidates compared with a baseline planner.

\end{abstract}


\section{Introduction} 
Navigation in visually impaired assistive settings poses unique challenges due to the lack of reliable visual cues for perceiving obstacles, environmental boundaries, and the motion of surrounding agents \cite{1}.
In practice, visually impaired assistance is often deployed in spatially constrained and structured environments, such as campus walkways, pedestrian corridors, and other low-speed, partially enclosed areas, where motion safety and comfort are strongly depend on the preacross consecutive trajectory segmentsof trajectory evolution \cite{28,29,30,36,37}.
In these scenarios, navigation typically follows predefined paths and requires frequent replanning in response to local interactions \cite{2}.
Under such conditions, abrupt changes in velocity or acceleration not only increase safety risks but also degrade motion comfort and perceptual consistency for users \cite{3,4}, making smooth and temporally coherent motion generation a critical requirement.

A practical solution in structured, low-speed scenarios commonly follows a two-stage procedure. Heuristic trajectory sampling is first employed to generate a trajectory cluster of candidate trajectory samples along a reference path, followed by local trajectory optimization to refine motion quality within the planning horizon \cite{5,6,31,38}.

In existing heuristic trajectory sampling approaches, candidate trajectories are commonly generated by independently sampling terminal configurations along a reference path \cite{32}.
For example, Ma et al.~\cite{7} employ multi-area endpoint sampling at fixed time intervals to construct candidate trajectories, which are subsequently evaluated and selected via cost-based optimization.
Similarly, in the context of trajectory prediction, Aydemir et al.~\cite{8} adopt endpoint-conditioned modeling, where trajectory hypotheses are parameterized with respect to independently modeled endpoints. However, under frequent replanning, independently modeled endpoints may yield irregular trajectory clusters and compromised segment continuity ~\cite{39}.

To address this limitation, an Endpoint Regulation mechanism is introduced at the trajectory sampling stage. By constraining terminal behavior within each trajectory cluster, endpoint regulation improves the stability and uniformity of trajectory sampling. As a result, trajectory samples exhibit more consistent segment boundary conditions, providing a reliable foundation for subsequent optimization. 

Existing local trajectory optimization methods commonly rely on smoothness penalties, particularly jerk minimization, to improve the smoothness of individual trajectory segments \cite{33}.
Representative studies have investigated jerk-constrained or minimum-jerk formulations in different application domains.
For instance, Lee et al.~\cite{9}, Ji et al.~\cite{10}, and Lozer et al.~\cite{11} studied jerk-aware or jerk-constrained local trajectory optimization for industrial manipulators, aiming to improve motion smoothness, energy efficiency, and tracking performance.
Similar jerk-based objectives have been adopted in vehicle motion planning to enhance ride comfort and safety during aggressive maneuvers \cite{12,13}, as well as in aerial robot trajectory generation to ensure high-order continuity \cite{14}. While effective at suppressing local oscillations, these penalties primarily act within a single trajectory segment and do not explicitly constrain motion transitions across consecutive trajectory segments \cite{40}. Consequently, velocity and acceleration may still change abruptly at segment boundaries under repeated replanning \cite{41}.

Furthermore, a momentum-aware local trajectory optimization formulation is proposed. The proposed formulation explicitly enforces momentum consistency across consecutive trajectory segments, directly regulating motion evolution at segment boundaries. This design strengthens segment continuity under frequent replanning and leads to smoother and safer motion behavior in structured, low-speed navigation scenarios. 

Built upon heuristic trajectory sampling, this work presents a unified trajectory generation approach that jointly addresses endpoint stability and motion consistency across consecutive trajectory segments. The proposed method is evaluated through three groups of experiments in representative structured, low-speed navigation scenarios, assessing trajectory segment smoothness, trajectory sampling stability and uniformity, and overall feasibility and safety. 

The main contributions of this paper are summarized as follows: 
\begin{itemize}
    \item This paper presents a trajectory generation pipeline for visually impaired scenarios that explicitly addresses segment inconsistency under frequent replanning. By jointly integrating endpoint regulation at the sampling stage and momentum-aware dynamics in local optimization, the proposed pipeline ensures coherent state evolution across consecutive regeneration cycles. 
    \item An endpoint regulation mechanism is devised to stabilize terminal state distributions of sampled trajectories for visually impaired scenarios, enforcing consistent boundary conditions for subsequent local optimization and mitigating segment-wise discontinuities under frequent replanning. 
    \item A momentum-aware dynamics objective is introduced for local trajectory optimization to meet the smoothness and comfort requirements of visually impaired, explicitly regularizing velocity and acceleration evolution across replanning cycles and suppressing abrupt state transitions at segment boundaries and enhances overall trajectory smoothness and temporal consistency. 
\end{itemize}

\section{Heuristic Trajectory Generation with Endpoint Regulation}
This section addresses trajectory inconsistency under frequent replanning and introduces an endpoint regulation mechanism to improve terminal coherence across sampled trajectories  for visually impaired
scenarios.

Trajectory generation and regulation are performed in the Frenet coordinate system, where agent motion is represented relative to a given reference path \cite{15,16}. 
Let $\boldsymbol{r}(s)$ denote the reference path parameterized by arc length $s$, which defines the local tangent and normal directions.
The Cartesian position of the agent is expressed as
\begin{equation}
\boldsymbol{x}(s,d) = \boldsymbol{r}(s) + d\,\boldsymbol{n}_c(s),
\end{equation}
where $d$ denotes the lateral offset from the reference path and $\boldsymbol{n}_c(s)$ is the unit normal vector.

The Frenet state is defined as
\begin{equation}
\boldsymbol{\xi}(t) =
\big[s,\dot{s},\ddot{s},d,\dot{d},\ddot{d}\big]^\top,
\end{equation}
which jointly characterizes longitudinal and lateral motion and their derivatives.
Throughout this work, the Frenet frame serves as the primary representation for trajectory sampling and regulation in visually impaired scenarios.

Given a reference path, heuristic trajectory samples are generated in the Frenet frame by decoupling longitudinal and lateral motions \cite{17,18,42}.
The longitudinal motion is parameterized as a quintic polynomial in time,
\begin{equation}
s(t) = \sum_{i=0}^{5} a_i t^i,
\label{eq:quintic_long}
\end{equation}
while the lateral motion is expressed with respect to the longitudinal displacement,
\begin{equation}
d(s) = \sum_{i=0}^{5} b_i s^i.
\label{eq:quintic_lat}
\end{equation}
Polynomial coefficients are determined by boundary conditions at the beginning and end of the planning horizon, ensuring continuity of position, velocity, and acceleration within each trajectory sample.

A trajectory cluster is formed by sampling multiple terminal configurations with respect to longitudinal progress, lateral offset, and horizon length.
Although each trajectory sample is smooth within its planning horizon, unconstrained terminal states may lead to excessive dispersion inside the trajectory cluster \cite{19,20,43}, producing poorly conditioned boundary states and weakening temporal compatibility for subsequent refinement, which is particularly problematic in visually impaired scenarios requiring frequent replanning.

To address this issue, an endpoint regulation mechanism is introduced to constrain terminal behavior across the trajectory cluster under frequent replanning conditions in visually impaired scenarios.
Rather than enforcing endpoint equality, the proposed regulation limits terminal deviation with respect to a reference trajectory sample, thereby improving boundary consistency while preserving sampling flexibility, which is essential for stable motion guidance in visually impaired scenarios.

\begin{figure*}[htbp]
\centering
\begin{tikzpicture}[x=1cm,y=1cm,>=Latex, font=\small]

% ---------- Common styles ----------
\tikzset{
  axis/.style={->, line width=0.6pt},
  traj/.style={line width=0.5pt, opacity=0.85},
  endpt/.style={circle, fill=black, inner sep=1.2pt},
  endpt2/.style={circle, fill=black, inner sep=1.0pt, opacity=0.9},
  refpt/.style={circle, fill=black, inner sep=2.2pt},
  bound/.style={dashed, line width=0.7pt},
  label/.style={font=\small}
}

% ---------- (a) Without endpoint regulation ----------
\begin{scope}[shift={(0,0)}]
  % Axes
  \draw[axis] (0,0) -- (5.2,0) node[below] {$s$};
  \draw[axis] (0,0) -- (0,3.2) node[left] {$d$};

  % Reference path (centerline)
  \draw[line width=0.9pt] (0.2,1.6) -- (5.0,1.6);
  \node[label] at (4.05,1.95) {reference path};

  % Sampled trajectories (schematic) + terminal states (irregular spacing)
    \foreach \y in {0.55,0.72,1.18,1.26,1.83,2.05,2.62,2.92} {
      \draw[traj] (0.3,1.6) .. controls (2.1,1.6) and (3.2,\y) .. (4.6,\y);
      \node[endpt] at (4.6,\y) {};
    }

  % Subfigure label
  \node[label] at (2.5,-0.55) {(a) Without endpoint regulation};

  % Note
  \node[label, align=center] at (2.55,3.05) {irregularly spaced\\ terminal states};
\end{scope}

% ---------- (b) With endpoint regulation ----------
\begin{scope}[shift={(7.0,0)}]
  % Axes
  \draw[axis] (0,0) -- (5.2,0) node[below] {$s$};
  \draw[axis] (0,0) -- (0,3.2) node[left] {$d$};

  % Reference path (centerline)
  \draw[line width=0.9pt] (0.2,1.6) -- (5.0,1.6);
  \node[label] at (4.05,1.95) {reference path};

  % Reference terminal state
  \node[refpt] (ref) at (4.6,1.6) {};
  \node[label, anchor=west] at (4.75,1.6) {reference terminal state};

  % Regulated deviation bound (ellipse)
  \draw[bound] (4.6,1.7) ellipse [x radius=0.80, y radius=1.25];

  \node[label, anchor=south] at (4.6,2.25) {regulated deviation bound};

  % Sampled trajectories (schematic, regulated endpoints) + terminal states (more uniform spacing)
\foreach \y in {0.55,0.77,0.99,1.21,1.43,1.65,1.87,2.09,2.31,2.53,2.75,2.92} {
  \draw[traj] (0.3,1.6) .. controls (2.1,1.6) and (3.2,\y) .. (4.6,\y);
  \node[endpt2] at (4.6,\y) {};
}

  % Subfigure label
  \node[label] at (2.5,-0.55) {(b) With endpoint regulation};

  % Note
  \node[label, align=center] at (2.55,3.05) {constrained\\ terminal states};
\end{scope}

\end{tikzpicture}
\caption{Schematic illustration of endpoint regulation in trajectory sampling.
(a) Unconstrained sampling yields dispersed terminal states within the trajectory cluster.
(b) Endpoint regulation constrains terminal deviation with respect to a reference sample, producing well-conditioned boundary states for subsequent local optimization.}
\label{fig:ep_reg}
\end{figure*}
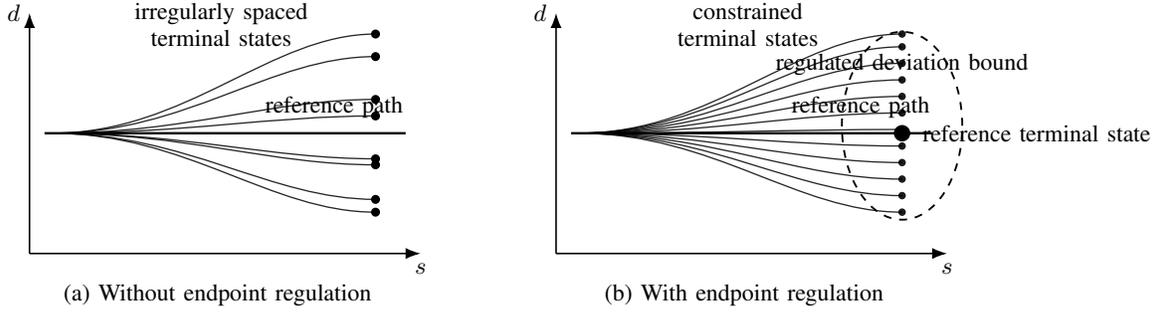

Fig.~\ref{fig:ep_reg} provides an intuitive illustration of the proposed endpoint regulation mechanism.
As shown in Fig.~\ref{fig:ep_reg}(a), without explicit regulation, terminal states generated by independent trajectory sampling may exhibit irregular spacing and heterogeneous dispersion within the trajectory cluster. Although all terminal states satisfy basic feasibility constraints, their uneven distribution leads to poorly conditioned boundary states, which complicates subsequent local refinement and degrades temporal compatibility across replanning cycles.

In contrast, Fig.~\ref{fig:ep_reg}(b) illustrates the effect of endpoint regulation, where terminal deviations are softly constrained with respect to a reference sample through a weighted regularization term. This mechanism reshapes the terminal-state distribution by reducing excessive dispersion and enforcing a more structured spacing pattern, while still allowing the overall extent of the distribution to be adaptively adjusted via the weighting matrix and spacing threshold. As a result, endpoint regulation yields a better-conditioned set of terminal states that is more suitable for downstream momentum-aware local optimization, particularly under frequent replanning conditions encountered in visually impaired assistive navigation.

Let the Frenet state of a trajectory sample be
$\boldsymbol{\xi}(t) = [s,\dot{s},\ddot{s},d,\dot{d},\ddot{d}]^\top$.
For a trajectory sample terminating at time $t_\tau$, the regulated terminal deviation is defined as
\begin{equation}
\mathcal{R}_{\mathrm{ep}}\!\left(\boldsymbol{\xi}(t_\tau)\right)
=
\left\|
\boldsymbol{W}_{\mathrm{ep}}
\left(
\boldsymbol{\eta}_i(t_\tau)
-
\boldsymbol{\eta}_{\mathrm{ref}}(t_\tau)
\right)
\right\|^2,
\label{eq:endpoint_reg_energy}
\end{equation}
where
$\boldsymbol{\eta}(t_\tau) = [\dot{s},\ddot{s},\dot{d},\ddot{d}]^\top$
collects terminal velocity and acceleration components,
$\boldsymbol{\eta}_{\mathrm{ref}}(t_\tau)$ denotes the terminal state of a reference trajectory sample within the same trajectory cluster,
and $\boldsymbol{W}_{\mathrm{ep}}$ is a diagonal weighting matrix.

This regulation suppresses excessive terminal variation across trajectory samples and yields a more coherent trajectory cluster at the planning boundary, which is particularly important for maintaining predictable and comfortable motion behavior in visually impaired assistive scenarios, while retaining sufficient diversity for downstream optimization.
To further avoid over-dispersion of terminal states, a minimal spacing constraint is imposed to regulate endpoint distribution within the trajectory cluster:
\begin{equation}
\left\|
\boldsymbol{\xi}_i(t_\tau)
-
\boldsymbol{\xi}_{i-1}(t_\tau)
\right\|
\le
\Delta_0,
\label{eq:endpoint_spacing}
\end{equation}
where $\Delta_0$ is a nominal spacing threshold.
This constraint prevents degenerate clustering of trajectory samples and promotes balanced coverage of the local planning space.

Through endpoint regulation, the trajectory sampling process produces a trajectory cluster with well-conditioned boundary states.
These regulated trajectory samples provide stable terminal conditions for momentum-aware local trajectory optimization.

Through endpoint regulation, the trajectory sampling process produces a trajectory cluster with well-conditioned boundary states, effectively mitigating segment-wise inconsistency induced by independent trajectory sampling, which is crucial for maintaining stable and predictable motion behavior under frequent replanning in visually impaired scenarios.
These regulated trajectory samples therefore provide reliable terminal conditions for momentum-aware local trajectory optimization.

\section{Momentum-aware Local Trajectory Optimization}

\subsection{Momentum Consistency Across Trajectory Segments}

Heuristic trajectory generation methods commonly construct trajectories by connecting multiple locally generated segments through endpoint constraints.
While geometric continuity is preserved, the temporal evolution of kinematic states across adjacent segments is not explicitly regulated.
Previous studies have reported that such formulations may lead to inconsistent state transitions near connection points, including abrupt changes in velocity or acceleration \cite{21,22}, which are particularly undesirable for visually impaired assistive scenarios due to their impact on user comfort and motion predictability.

\begin{figure}[htbp]
\centerline{\includegraphics[width =.45\textwidth]{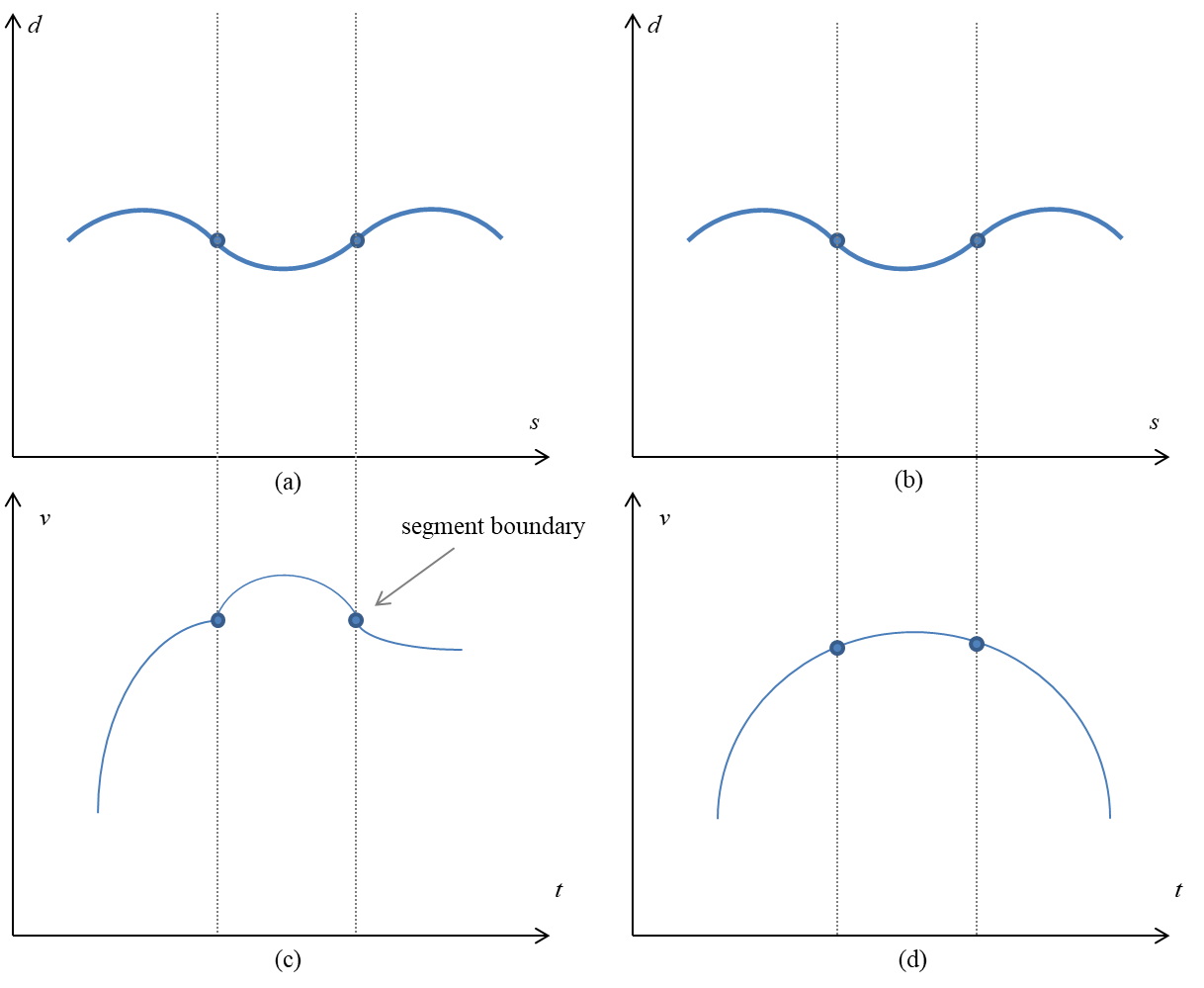}}
\caption{Illustration of momentum inconsistency in heuristic trajectory generation.
(a) Spatial trajectory composed of independently optimized segments.
(b) Spatial trajectory with momentum-aware regulation, showing similar geometric layout.
(c) Velocity profile without momentum consistency, exhibiting a slope discontinuity at the segment boundary.
(d) Velocity profile with enforced momentum consistency, yielding smooth temporal evolution.}
\label{fig1}
\end{figure}

Fig.~\ref{fig1}(a) presents a spatial trajectory composed of consecutive segments, where position continuity is maintained at the connection point.
However, the corresponding state evolution is not directly observable from the spatial representation.
By incorporating momentum-aware regulation at segment transitions, the spatial trajectory shown in Fig.~\ref{fig1}(b) remains geometrically similar, indicating that the proposed regulation does not alter the overall path shape but modifies the underlying state transitions. The inconsistency becomes evident in the temporal domain.
As illustrated in Fig.~\ref{fig1}(c), the velocity profile associated with unconstrained segment connections is continuous in magnitude but exhibits a slope discontinuity at the connection point, implying an abrupt change in acceleration.
In contrast, Fig.~\ref{fig1}(d) shows that the proposed momentum-aware dynamics yield a smooth state transition, where both the velocity and its derivative remain continuous across the connection, supporting predictable and comfortable motion behavior required in visually impaired scenarios.

These observations demonstrate that endpoint continuity in the spatial domain alone is insufficient to ensure consistent state evolution across trajectory segments \cite{44}.
In particular, conventional smoothness objectives, such as jerk minimization, mainly penalize local high-order derivatives within individual trajectory segments and do not explicitly enforce momentum consistency across segment transitions \cite{23,24}.
As a result, continuity in momentum evolution cannot be guaranteed, and state inconsistencies may still occur at segment boundaries even when geometric smoothness is preserved, which undermines motion stability for visually impaired scenarios under frequent replanning.

To address this limitation, a momentum-aware dynamics objective is incorporated into local trajectory generation to explicitly regularize state consistency across consecutive regeneration cycles and suppress discontinuities at segment boundaries, in accordance with the strict smoothness and comfort requirements for visually impaired scenarios. 
By directly constraining momentum evolution rather than relying solely on local smoothness penalties, the proposed formulation provides a principled mechanism to bridge the gap between spatial continuity and temporally consistent state transitions, forming a core component of trajectory generation for visually impaired scenarios.

\subsection{Momentum-Aware Formulation}

To characterize the temporal evolution of states during trajectory generation under frequent replanning, a continuous-time representation of the agent dynamics is adopted to support smooth and predictable motion behavior required in visually impaired scenarios. 
This formulation provides a unified description of longitudinal and lateral motion evolution and serves as the modeling basis for the subsequent momentum-aware optimization objective.

Specifically, the state dynamics and the linear observation structure are expressed as:
\begin{equation}
\begin{aligned}
\dot{\boldsymbol{\xi}}(t) & = \mathcal{A}\,\boldsymbol{\xi}(t) + \mathcal{B}\,\boldsymbol{u}(t), \\
\boldsymbol{y}(t) & = \mathcal{C}\,\boldsymbol{\xi}(t) + \mathcal{D}\,\boldsymbol{u}(t)
\in \mathcal{R}(t),
\end{aligned}
\end{equation}
where $\mathcal{A}$ and $\mathcal{B}$ define a linear kinematic evolution model, and $\mathcal{C}$ and $\mathcal{D}$ specify a linear observation structure.

In this formulation, $\boldsymbol{\xi}(t)=[s,\dot{s},\ddot{s},d,\dot{d},\ddot{d}]^\top$ collects the longitudinal and lateral motion states in the Frenet frame, while $\boldsymbol{u}(t)$ denotes the control input associated with the change of the system.
It should be noted that the above model describes kinematic motion dynamics for planning and regularization purposes, rather than a full physical dynamics model. $\boldsymbol{y}(t)$ denotes the observable motion output used for feasibility evaluation, and $\mathcal{R}(t)$ defines a time-varying admissible region determined by environmental constraints and kinematic limits.

Let $v_i(t)$ represent the velocity in the local motion frame.
the following optimization objective is defined over a finite horizon $[t_0,t_\tau]$:
\begin{equation}
\mathcal{J}_i
=
\int_{t_0}^{t_\tau} L_i(t)\,\mathrm{d}t
+
\lambda_{\mathrm{ep}}
\,
\mathcal{R}_{\mathrm{ep}}\!\left(\boldsymbol{\xi}(t_\tau)\right),
\label{eq:mto_obj}
\end{equation}
where $\mathcal{R}_{\mathrm{ep}}(\cdot)$ is a terminal regularizer inherited from endpoint regulation, ensuring compatibility with segment continuity constraints.

\paragraph{Momentum-Aware Lagrangian.}
The Lagrangian is constructed to explicitly enforce momentum consistency across consecutive trajectory segments:
\begin{equation}
\begin{aligned}
L_i(t)
=
&\;\underbrace{\frac{1}{2} m_i \|v_i(t)\|^2}_{\textbf{kinetic energy}}
-
\underbrace{\boldsymbol{F}_{\text{ext}}(t)\!\cdot\! v_i(t)}_{\textbf{external momentum modulation}} \\
&+
\underbrace{\lambda_s \|\dot{v}_i(t)\|^2}_{\textbf{momentum-change suppression}}
+
\underbrace{\lambda_u\,\Psi_u(t)}_{\textbf{robustness regularization}},
\end{aligned}
\label{eq:mto_lagrangian_compact}
\end{equation}
where $m_i$ is the agent mass and $\lambda_s,\lambda_u\ge0$ are weighting coefficients.

Unlike jerk-based penalties that act locally within a trajectory segment, this formulation directly penalizes inconsistent momentum evolution and thus suppresses discontinuities at segment boundaries \cite{25}.

\paragraph{External modulation abstraction.}
The modulation term $\boldsymbol{F}_{\text{ext}}(t)$ is introduced as an abstract momentum-shaping input in the Frenet frame, rather than a physical force.
Since the state $\boldsymbol{\xi}(t)=[s,\dot{s},\ddot{s},d,\dot{d},\ddot{d}]^\top$ jointly describes longitudinal and lateral motion, $\boldsymbol{F}_{\text{ext}}(t)$ is defined as a two-dimensional vector acting on both components.
It is decomposed into a guidance-related part and an interaction-induced part:
\begin{equation}
\boldsymbol{F}_{\text{ext}}(t)
=
-\boldsymbol{F}_{\text{asst}}(t)
+
\boldsymbol{F}_{\text{int}}(t),
\qquad
\boldsymbol{F}_{\text{asst}},\boldsymbol{F}_{\text{int}}\in\mathbb{R}^2.
\label{eq:fext_decomp}
\end{equation}

$\boldsymbol{F}_{\text{asst}}(t)$ summarizes bounded assistive cues that bias the motion toward a stable pace and corridor adherence along the reference path, e.g.,
\begin{equation}
\boldsymbol{F}_{\mathrm{asst}}(t)
=
\begin{bmatrix}
f_s\!\big(\dot{s}(t),\,\beta(s(t))\big) \\
f_d\!\big(d(t),\dot{d}(t),\,\beta_(d(t))\big)
\end{bmatrix},
\qquad
\|\boldsymbol{F}_{\mathrm{asst}}(t)\|\le \bar{F}_{\mathrm{asst}} .
\label{eq:asst_generic}
\end{equation}

Here $\beta_(s)$ is a bounded descriptor of surface irregularity along the reference path.
Both $f_s(\cdot)$ and $f_d(\cdot)$ are smooth bounded shaping functions. $\beta_(s)$ captures bump-induced disturbances with speed-amplified effect implicitly reflected through $\dot{s}$, while the lateral boundary-related corrective tendency is encoded through the dependence on $d$ rather than introducing an additional boundary-activation term.

$\boldsymbol{F}_{\text{int}}(t)$ captures bounded influences induced by surrounding agents/obstacles:
\begin{equation}
\boldsymbol{F}_{\mathrm{int}}(t)
=
\sum_{j\in\mathcal{N}_i}
\alpha_{ij}(t)\,\boldsymbol{n}_{ij}(t),
\label{eq:int_main}
\end{equation}
where
\begin{equation}
\begin{aligned}
\alpha_{ij}(t)
&=
\alpha\!\left(
r_{ij}(t),\,
\Delta v_{ij}(t)
\right),
\qquad
0 \le \alpha_{ij}(t) \le \bar{\alpha}.
\end{aligned}
\label{eq:int_compact_aligned}
\end{equation}

$r_{ij}(t)$ and $\Delta v_{ij}(t)$ denote the relative distance and relative speed magnitude, respectively.
The interaction intensity $\alpha_{ij}(\cdot)$ is a smooth bounded function that increases with decreasing distance and increasing relative speed, while the direction of interaction is fully encoded by the unit vector $\boldsymbol{n}_{ij}(t)$. This abstraction preserves interpretability by modeling repulsive interactions along the direction vector $\boldsymbol{n}_{ij}$, while avoiding over-specification of any particular social-force model.

Overall, the decomposition in \eqref{eq:fext_decomp} provides a lightweight interface to incorporate assistive guidance and interaction influences into the momentum-aware Lagrangian, while keeping the formulation agnostic to the exact realization of $\boldsymbol{F}_{\text{asst}}$ and $\boldsymbol{F}_{\text{int}}$.

\paragraph{Uncertainty-aware regularization.}
To promote conservative momentum evolution, a robustness term is introduced:
\begin{equation}
\Psi_u(t)
=
\operatorname{Tr}\!\left[\Sigma(t)\right],
\label{eq:uncertainty_proxy}
\end{equation}
where $\Sigma(t)$ is a generic uncertainty descriptor provided by the scenario representation.
This term serves as an auxiliary regularizer and does not alter the structure of the momentum constraint.

\subsection{Local Trajectory Optimization Procedure}

The optimal local trajectory is obtained by minimizing \eqref{eq:mto_obj}:
\begin{equation}
\boldsymbol{u}^*
=
\arg\min_{\boldsymbol{u}(\cdot)}\;
\mathcal{J}_i,
\label{eq:mto_u_opt}
\end{equation}
which yields the Euler--Lagrange conditions with respect to $\boldsymbol{\xi}(t)$:
\begin{equation}
\frac{\mathrm{d}}{\mathrm{d}t}
\Big(
\frac{\partial L}{\partial \dot{\boldsymbol{\xi}}}
\Big)
-
\frac{\partial L}{\partial \boldsymbol{\xi}}
+
\frac{\mathrm{d}^2}{\mathrm{d}t^2}
\Big(
\frac{\partial L}{\partial \ddot{\boldsymbol{\xi}}}
\Big)
=0.
\label{eq:EL}
\end{equation}
In practice, the resulting equations are solved using a discretized backtracking procedure along the Frenet horizon.
This momentum-aware dynamics improves momentum consistency across trajectory segments while preserving the efficiency of heuristic sampling, thereby resolving a limitation of conventional segment-wise planning and providing an enabling condition for stable, predictable motion execution in visually impaired scenarios.

\begin{figure*}[!t]
\centerline{\includegraphics[width =.99\textwidth]{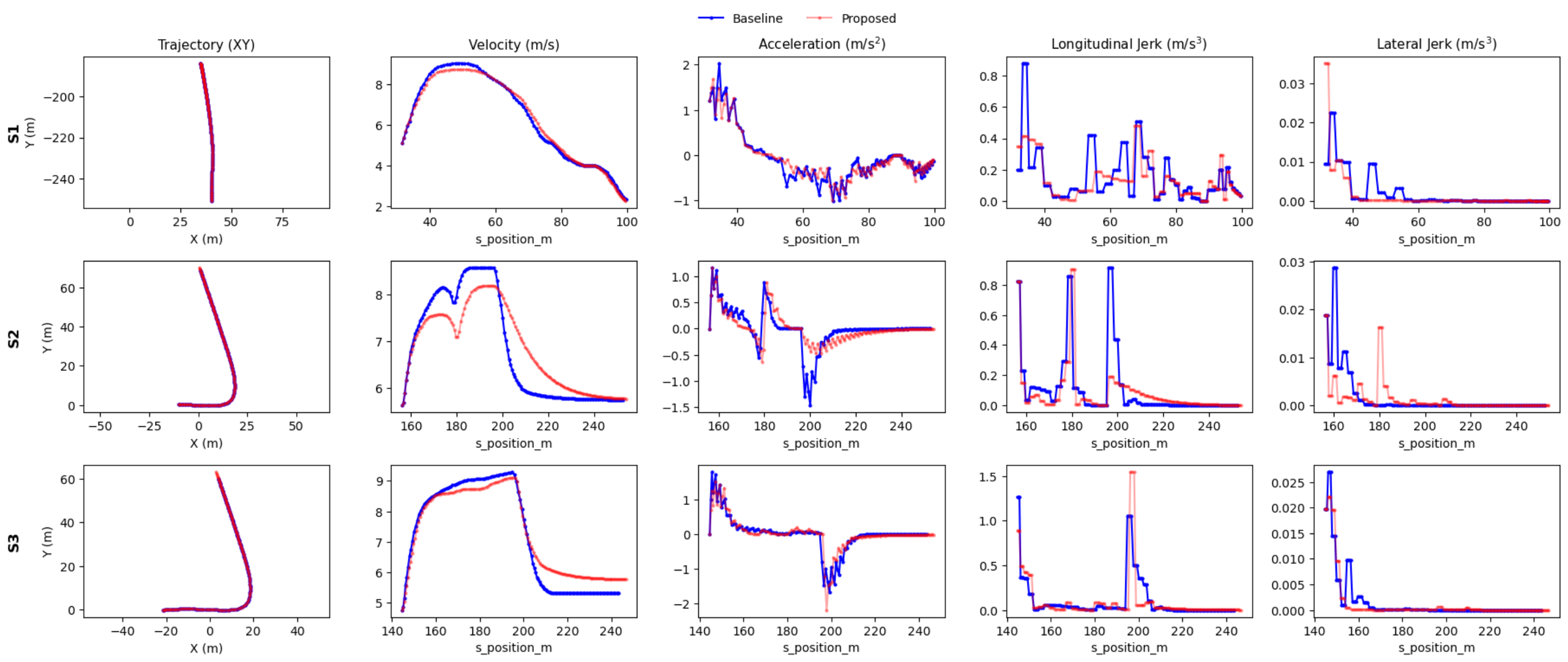}}
\caption{Velocity, acceleration, and jerk profiles along the longitudinal displacement across three low-speed scenarios.
The results compare the baseline and the proposed method, highlighting consistent trends in motion smoothness and jerk attenuation.}

\label{fig2345}
\end{figure*}
\section{Experimental Evaluation}

This section evaluates the proposed method for visually impaired scenarios.
All experiments are conducted in simulation and focus on assessing trajectory smoothness, endpoint regulation behavior, and overall feasibility and safety under consistent experimental conditions.

\subsection{Experimental Setup}
The experiments are conducted in three structured, low-speed navigation scenarios selected from the benchmark \cite{26,45}.
These scenarios are characterized by relatively low speed limits, predefined road geometries, and a limited set of interacting agents.
In addition, the selected scenarios exhibit spatially constrained and partially enclosed layouts, such as road segments with geometric boundaries, which together reflect conditions commonly encountered in visually impaired scenarios such as campus walkways and semi-enclosed areas. 

For each scenario, the reference path as well as the initial velocity and acceleration states are directly adopted from the scenario specification to ensure reproducibility.

The proposed method is compared with a baseline planner \cite{27} under identical scenario conditions.
Both methods follow the same trajectory generation and selection pipeline and share identical kinematic feasibility limits, including bounds on velocity, acceleration, and jerk.
Therefore, observed performance differences can be attributed to differences in the underlying planning formulations rather than to scenario settings or constraint configurations.

\subsection{Experiment 1: Local Smoothness Analysis}
The experiment first evaluate the baseline and the proposed method in three low-speed navigation scenarios under identical initial conditions and planning parameters, highlighting the systematic effect of momentum and endpoint regulation on motion smoothness. 

As summarized in Fig.~\ref{fig2345}, all three scenarios exhibit similar behavioral patterns, where S1, S2, and S3 correspond to three selected low-speed structured scenarios representative of visually impaired assistive navigation.
The baseline method consistently shows oscillatory velocity and spiky acceleration profiles, whereas the proposed approach yields smoother velocity transitions and suppresses acceleration peaks.
This trend is further reflected in the jerk profiles, where both longitudinal and lateral components present lower amplitudes and reduced high-frequency variations across all scenarios.

\begin{figure}[htbp]
\centerline{\includegraphics[width =.45\textwidth]{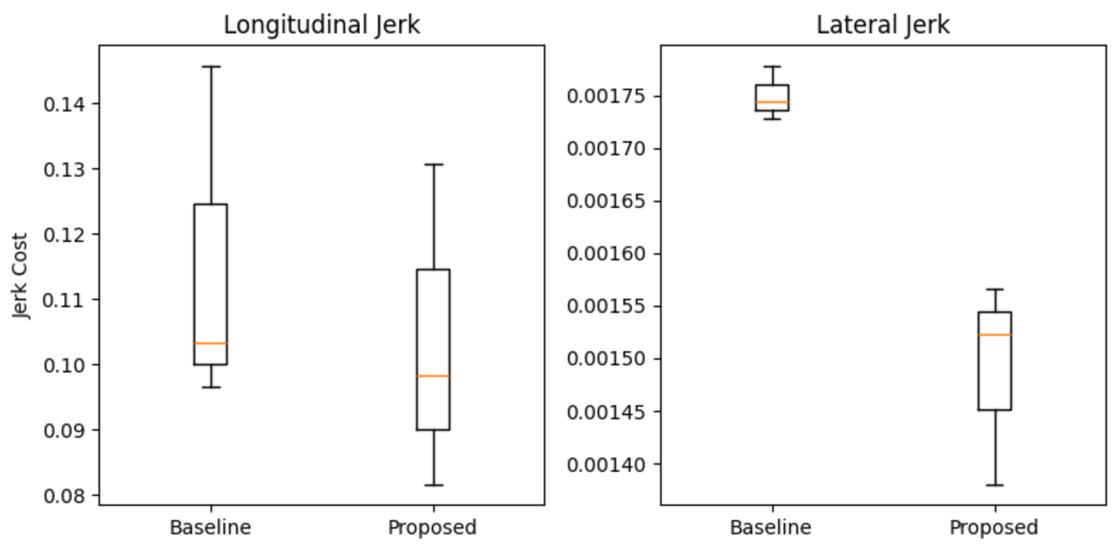}}
\caption{Boxplots of longitudinal and lateral jerk across three scenarios.
}
\label{fig6}
\end{figure}

The experiment further analyze jerk statistics across all three scenarios. As shown in Fig.~\ref{fig6}, the proposed method consistently yields lower median values and reduced dispersion for both longitudinal and lateral jerk compared with the baseline across all scenarios.

Together, these results is not limited to segment smoothness, but reflects enhanced continuity of kinematic states across consecutive replanning cycles, which is a critical requirement for stable guidance in visually impaired assistive navigation.
\subsection{Experiment 2: Endpoint Density Regulation}
This experiment evaluates the effect of endpoint regulation on the stability of trajectory sampling by analyzing the density and dispersion of sampled trajectory endpoints.

\begin{figure}
\centering
\begin{subfigure}{0.48\linewidth}
    \centerline{\includegraphics[width = .99\textwidth]{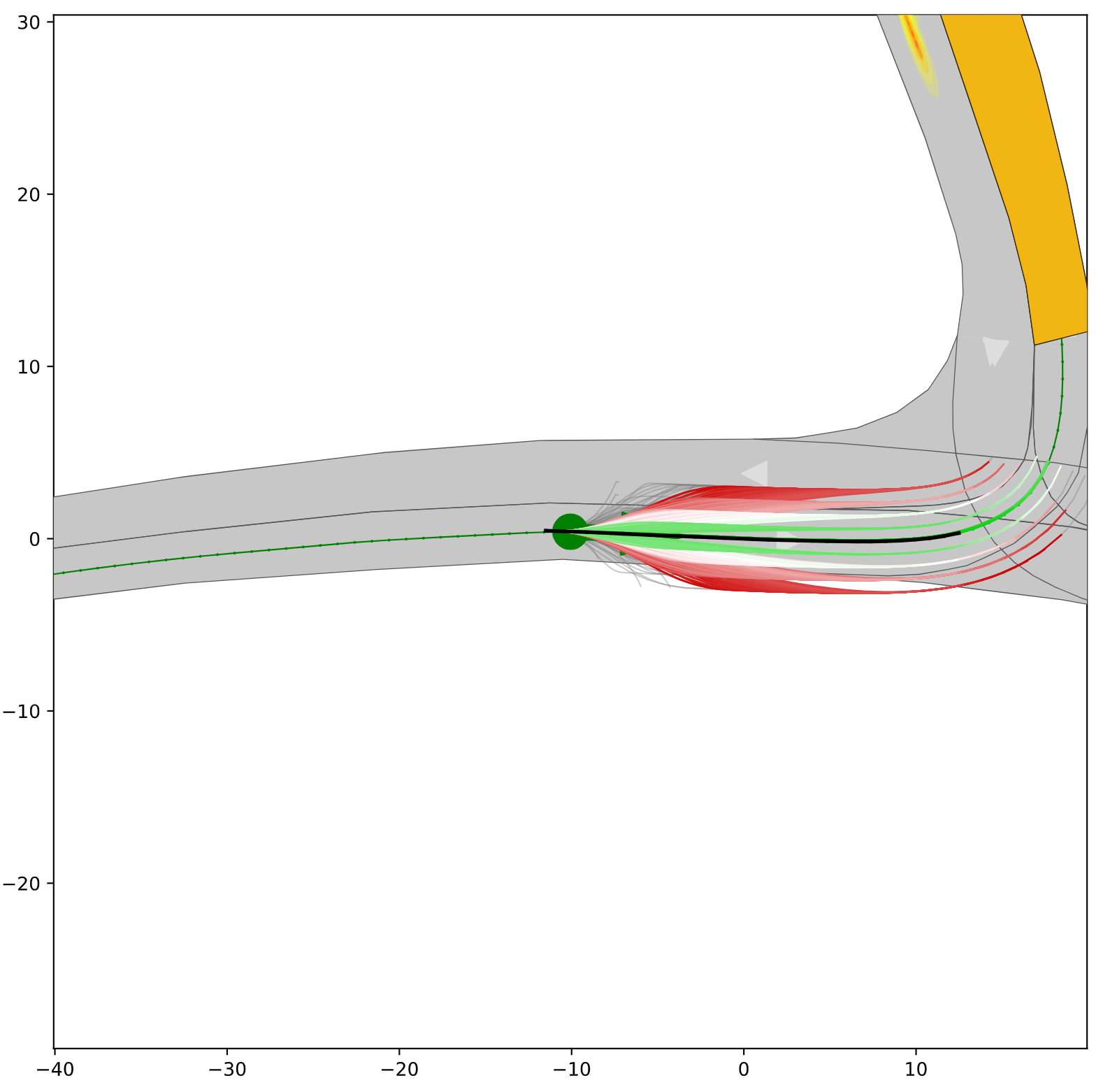}}
    \caption{}
    \label{fig10a}
\end{subfigure}
\begin{subfigure}{0.49\linewidth}
    \centerline{\includegraphics[width = .99\textwidth]{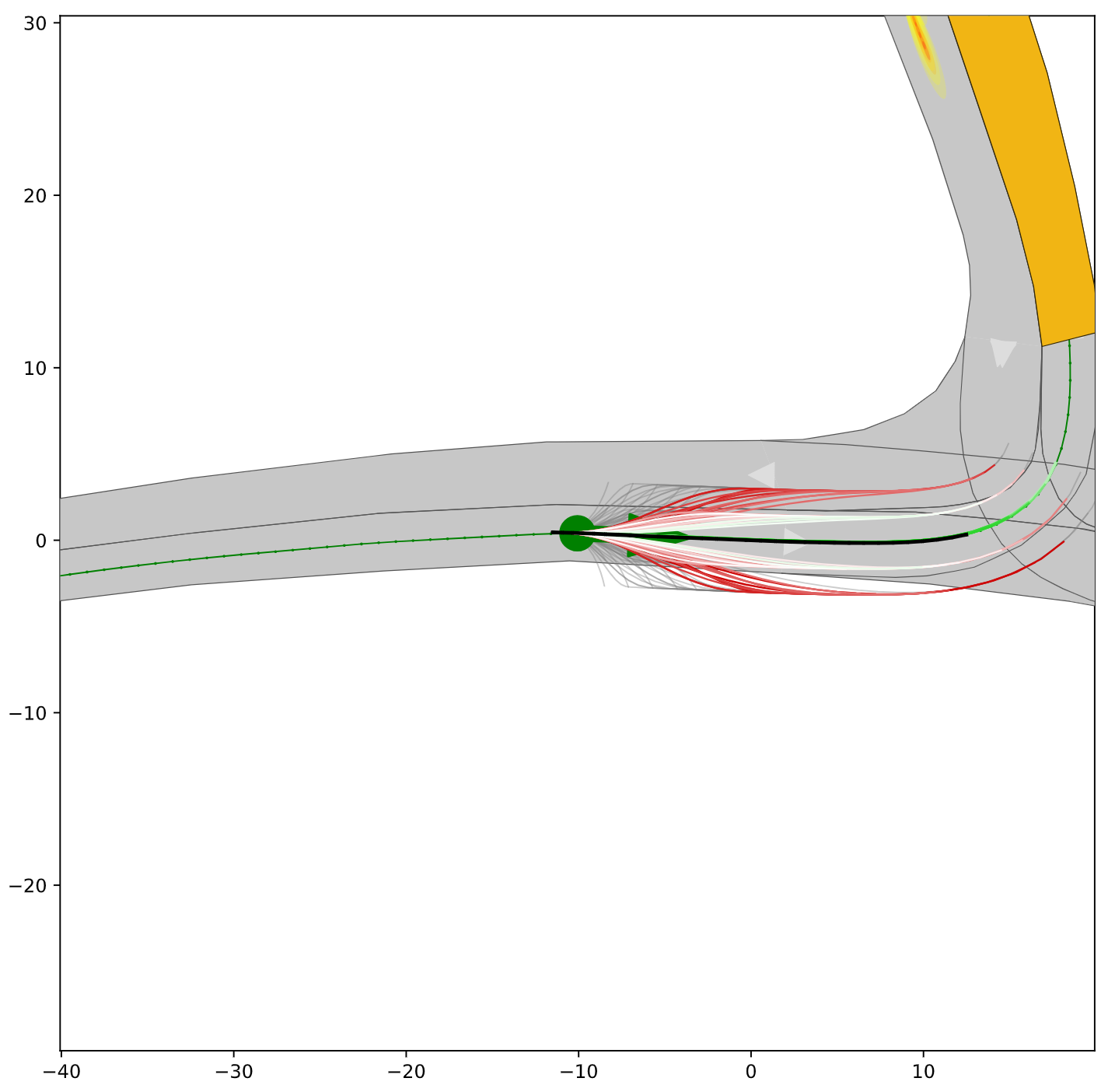}}
    \caption{}
    \label{fig10b}
\end{subfigure}
\caption{Comparison of trajectory cluster generation with and without endpoint regulation at a representative time step,
where (a) corresponds to the baseline and (b) shows the result with endpoint regulation.}
\label{fig10}
\end{figure}

Fig.~\ref{fig10} provides a qualitative comparison of the trajectory clusters generated by the baseline and the proposed method
at a representative time step in one scenario.
The figure illustrates differences in the spatial organization of the generated trajectories and the resulting endpoint distribution patterns. Compared with the baseline, the proposed method produces a more structured and evenly distributed trajectory cluster,
while maintaining comparable coverage of the reachable space.

This qualitative observation is consistent with the endpoint density statistics reported in Table~\ref{tab:endpoint_density}. For each of the three scenarios, endpoint statistics are computed based on the nearest-neighbor distance between sampled trajectory endpoints in the Frenet frame.
Table~\ref{tab:endpoint_density} summarizes the resulting density statistics for the baseline and the proposed method, where B and P denote the baseline and the proposed methods, respectively.

\begin{table}[!t]
\caption{Endpoint density statistics based on nearest-neighbor distance.}
\label{tab:endpoint_density}
\centering
\setlength{\tabcolsep}{5pt}
\renewcommand{\arraystretch}{1.1}
\begin{tabular}{lcccc}
\hline
Scenario & Method & Mean$\pm$Std & Min & Max \\
\hline
S1 & B \cite{27} & 0.556$\pm$0.214 & 0.232 & 0.907 \\
   & P & 0.556$\pm$\textbf{0.212} & 0.226 & 0.875 \\
S2 & B \cite{27} & 0.663$\pm$0.108 & 0.566 & 0.863 \\
   & P & 0.673$\pm$\textbf{0.084} & 0.566 & 0.818 \\
S3 & B \cite{27} & 0.678$\pm$0.165 & 0.480 & 0.926 \\
   & P & 0.700$\pm$\textbf{0.135} & 0.479 & 0.910 \\
\hline
Std of scenario means 
    & B \cite{27} & 0.066 & -- & -- \\
    & P & \textbf{0.077} & -- & -- \\
Avg. within-scenario Std 
    & B \cite{27} & 0.162 & -- & -- \\
    & P & 0.144 & -- & -- \\

\hline
\end{tabular}
\end{table}
Across all scenarios, the proposed method exhibits comparable mean nearest-neighbor distances while consistently reducing within-scenario dispersion, as reflected by lower standard deviations.
Although the absolute changes in mean distance are modest, these statistics indicate that endpoint regulation primarily affects the distribution regularity rather than shifting the overall sampling scale.

\begin{figure}[htbp]
\centerline{\includegraphics[width =.45\textwidth]{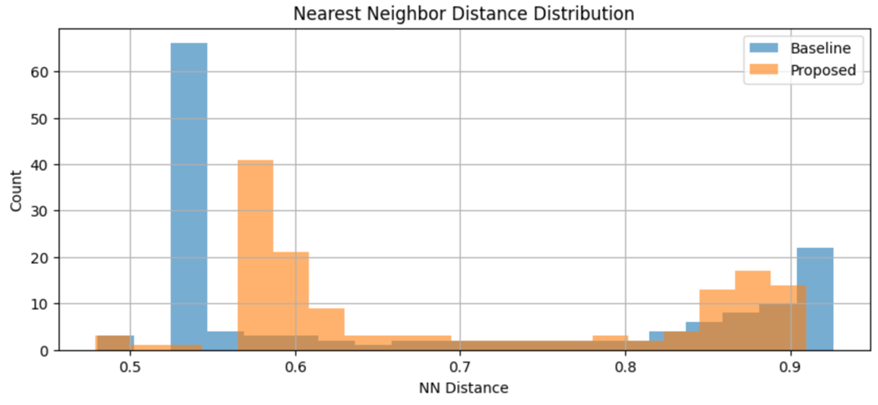}}
\caption{Nearest-neighbor distance distribution of trajectory endpoints in a representative scenario.}
\label{fig7}
\end{figure}

To provide an intuitive illustration of this effect, Fig.~\ref{fig7} visualizes the nearest-neighbor distance distribution for a representative scenario.
As shown in the figure, the proposed method yields a more evenly distributed nearest-neighbor distance profile, avoiding excessive concentration at specific distance values observed in the baseline.

Overall, the statistical results and representative visualization consistently show that endpoint regulation improves the stability and uniformity of trajectory sampling across different scenarios. This improved regularity of endpoint distribution leads to better-conditioned boundary states for subsequent local optimization and trajectory selection, reducing sensitivity to endpoint outliers under frequent replanning for visually impaired scenarios.

\subsection{Experiment 3: Feasibility and Safety}
The third experiment evaluates feasibility and safety characteristics of the generated trajectory candidates across three scenarios, which are critical for reliable planning in visually impaired assistive navigation.

At each planning cycle, each trajectory in the sampling cluster is evaluated against feasibility criteria, including kinematic constraints, and classified as feasible or infeasible.
The resulting statistics are aggregated across scenarios for comparison.
\begin{figure}
\centering
\begin{subfigure}{0.59\linewidth}
    \centerline{\includegraphics[width = .99\textwidth]{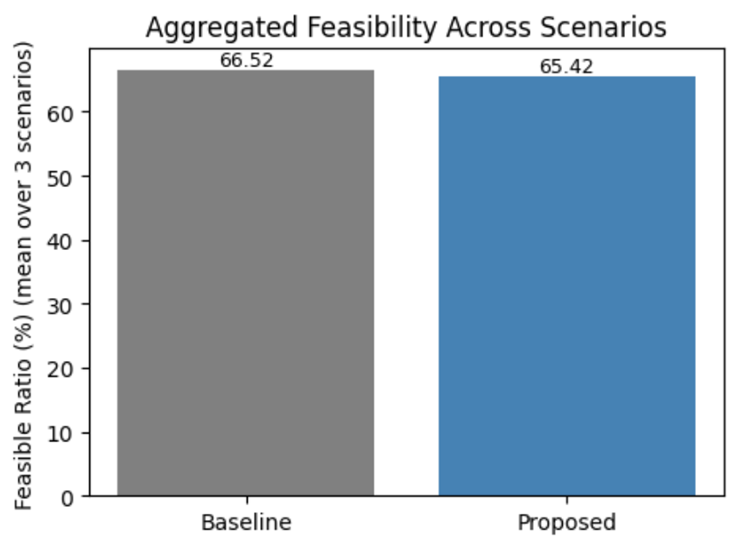}}
    \caption{}
    \label{fig9a}
\end{subfigure}

\begin{subfigure}{0.99\linewidth}
    \centerline{\includegraphics[width = .99\textwidth]{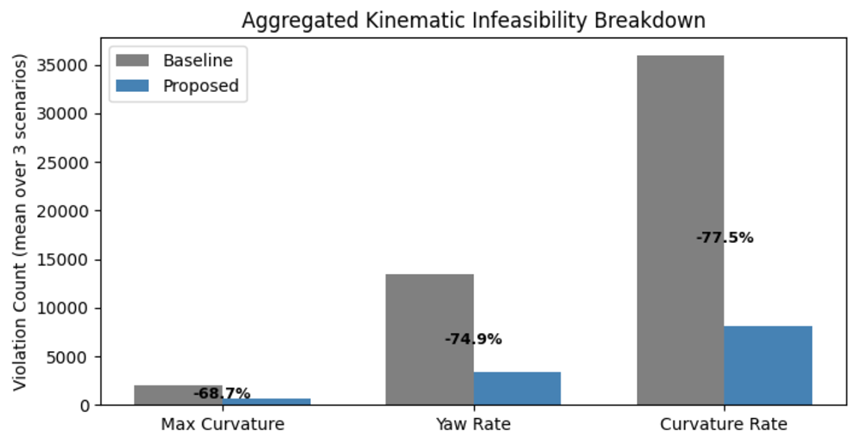}}
    \caption{}
    \label{fig9b}
\end{subfigure}
\caption{Bar charts summarizing feasibility and infeasibility statistics across three scenarios, where (a) reports the overall feasibility ratio and (b) compares the rates of infeasibility due to maximum curvature, yaw rate, and curvature rate constraints.}
\label{fig9}
\end{figure}
As shown in Fig.~\ref{fig9}, the proposed method maintains an overall feasibility ratio comparable to that of the baseline, while substantially reducing infeasibility rates associated with safety-critical constraints, including maximum curvature \cite{34}, yaw rate \cite{35}, and curvature rate.
These constraint violations correspond to trajectories with excessive curvature or abrupt directional changes, which are undesirable in assistive navigation scenarios.

By reducing the proportion of such unsafe candidates without decreasing the overall feasibility level, the proposed method yields a trajectory cluster that is both safer and more effective for downstream trajectory selection, indicating that endpoint regulation and momentum-aware optimization jointly improve safety without sacrificing sampling efficiency.

\section{Conclusion}
This paper investigated trajectory generation under frequent replanning for visually impaired assistive scenarios.
A trajectory generation approach is proposed that integrates endpoint regulation during sampling construction and momentum-aware dynamics during local trajectory optimization to improve segment consistency across replanning cycles. 
By incorporating endpoint regulation during trajectory sampling and explicitly constraining momentum evolution in the local optimization stage, the proposed method effectively reduces segment-wise discontinuities without altering the overall planning pipeline.

Experimental results across multiple representative scenarios show that the proposed method suppresses abrupt acceleration and jerk, produces more stable endpoint distributions, and reduces the number of infeasible trajectory candidates compared with a baseline planner.
These improvements lead to smoother, more predictable, and safer motion behavior, which is particularly important for visually impaired assistive navigation in structured, low-speed environments.

Future work will explore extending the method to more complex and densely populated scenarios and investigating adaptive parameter strategies to improve robustness under varying environmental conditions.

% \section*{Acknowledgment}

\bibliographystyle{IEEEtran}
\bibliography{references}

% \begin{thebibliography}{00}
% \bibitem{b1} G. Eason, B. Noble, and I. N. Sneddon, ``On certain integrals of Lipschitz-Hankel type involving products of Bessel functions,'' Phil. Trans. Roy. Soc. London, vol. A247, pp. 529--551, April 1955.
% \bibitem{b2} J. Clerk Maxwell, A Treatise on Electricity and Magnetism, 3rd ed., vol. 2. Oxford: Clarendon, 1892, pp.68--73.
% \bibitem{b3} I. S. Jacobs and C. P. Bean, ``Fine particles, thin films and exchange anisotropy,'' in Magnetism, vol. III, G. T. Rado and H. Suhl, Eds. New York: Academic, 1963, pp. 271--350.
% \bibitem{b4} K. Elissa, ``Title of paper if known,'' unpublished.
% \bibitem{b5} R. Nicole, ``Title of paper with only first word capitalized,'' J. Name Stand. Abbrev., in press.
% \bibitem{b6} Y. Yorozu, M. Hirano, K. Oka, and Y. Tagawa, ``Electron spectroscopy studies on magneto-optical media and plastic substrate interface,'' IEEE Transl. J. Magn. Japan, vol. 2, pp. 740--741, August 1987 [Digests 9th Annual Conf. Magnetics Japan, p. 301, 1982].
% \bibitem{b7} M. Young, The Technical Writer's Handbook. Mill Valley, CA: University Science, 1989.
% \end{thebibliography}
\vspace{12pt}
\color{red}
% IEEE conference templates contain guidance text for composing and formatting conference papers. Please ensure that all template text is removed from your conference paper prior to submission to the conference. Failure to remove the template text from your paper may result in your paper not being published.

\end{document}